# WhiteCon: Semi-Supervised Domain Adaptation Regression Through Whitening Transform and Dual Consistency

Se Jin Sim[0009–0000–6028–1690] and Seoung Bum Kim[0000–0002–2205–8516]

School of Industrial and Management Engineering
Korea University, Seoul, Republic of Korea
{ssj259, sbkim1}@korea.ac.kr

**Abstract.** Domain adaptation is crucial for addressing distributional shifts that degrade model performance across domains. While most existing research has centered on classification, semi-supervised domain adaptation regression (SSDAR) for continuous-output tasks remains largely unexplored, particularly in practical scenarios with limited labeled target data. To address this gap, we propose semi-supervised domain adaptation regression through whitening transform and dual consistency (WhiteCon), which combines domain-specific whitening transform (DWT) and dual consistency regularization to enhance training stability and domain adaptation. DWT reduces the variance of the model parameters by transforming the feature covariance matrix into an identity matrix, thus stabilizing training under ordinary least squares assumptions. In addition, variance consistency regularization, as part of dual consistency regularization, aligns the variances of weak, strong, and mixup-augmented features to improve resilience against augmentation-induced perturbations. Empirical evaluations on various benchmark datasets under SSDAR settings demonstrate that the proposed WhiteCon achieves state-of-the-art performance compared to existing methods, effectively addressing domain shifts in regression tasks. The code for WhiteCon is available at https://github.com/sejin-sim/WhiteCon.

**Keywords:** semi-supervised domain adaptation, regression, feature whitening, consistency regularization, variance alignment.

## 1 Introduction

Domain adaptation is a critical area of research that addresses performance degradation caused by distribution shifts between different domains. This topic has recently gained substantial attention in various fields, such as computer vision, natural language processing, and medical data analysis [1-3]. When a model trained on a source domain is applied to a target domain, significant differences in data distributions across domains can degrade the model's generalization performance [4, 5]. This issue is particularly challenging when labeled data are limited, emphasizing the need for domain adaptation that can reduce distribution gaps to ensure reliable performance in diverse settings [6].

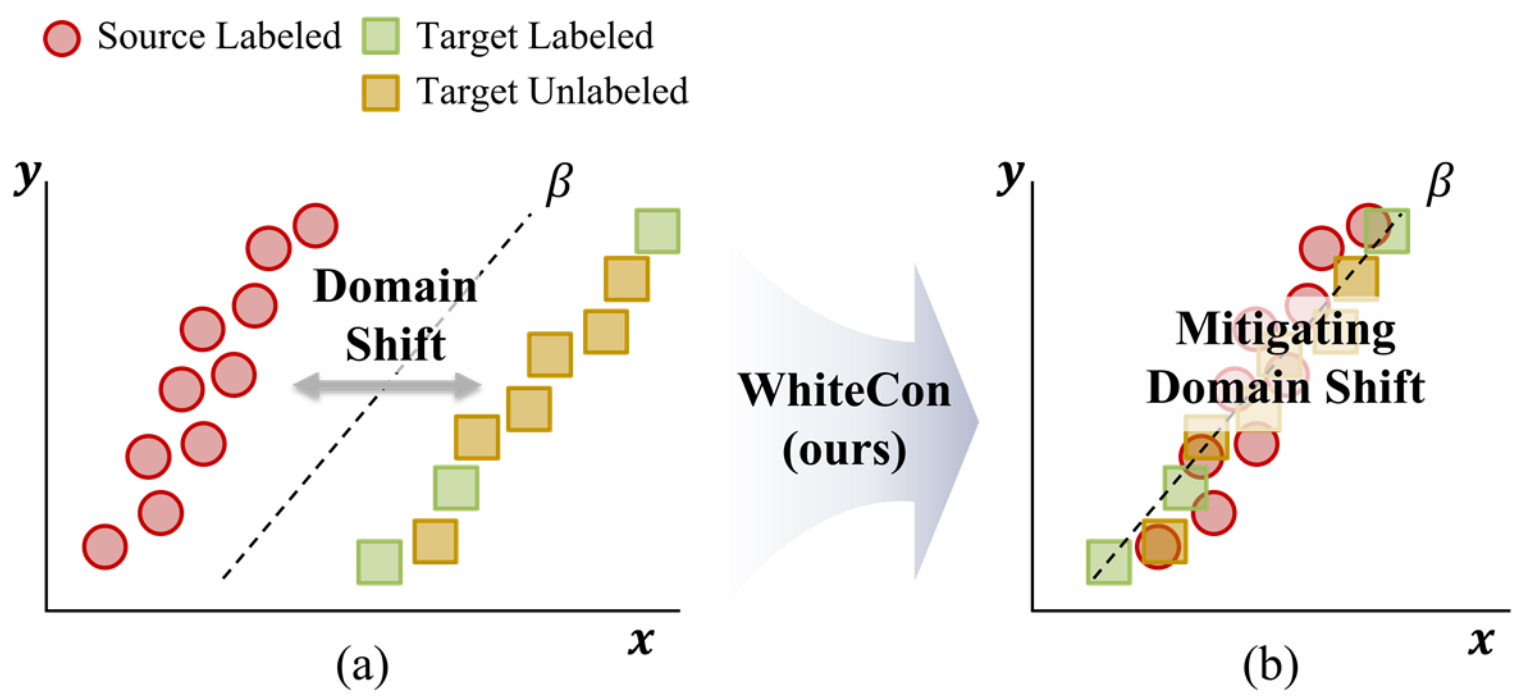

**Fig. 1.** Illustration of the OLS problem in SSDAR using WhiteCon. $\beta$ represents the parameters of the regressor. (a) Domain shift between source and target data under OLS. (b) WhiteCon mitigates the domain shift in SSDAR under OLS.

Most existing research on domain adaptation has focused on classification problems, leading to the development of various methods for reducing domain differences. Some methods, such as statistical moment alignment, aim to minimize distributional discrepancies [7, 8]. Another effective strategy, adversarial training, uses a minimax game between feature extractors and domain classifiers to reduce domain discrepancies, yielding promising results [9, 10]. Although some classification methods can be extended to regression, their reliance on class probability distributions is poorly suited to continuous outputs, creating a significant challenge for regression-based domain adaptation [11].

In response to these challenges, unsupervised domain adaptation regression (UDAR) methods have been proposed [12, 13]. UDAR methods aim to reduce domain discrepancies by using only labeled source data and unlabeled target data, making them particularly useful when labeled target data is scarce. Several methods, including singular value decomposition and distribution matching in feature space, have been used to align source and target domains [11, 14]. However, even small amounts of labeled target data significantly improve performance and are often feasible to acquire [13, 15].

As a result, semi-supervised domain adaptation (SSDA) integrates small amounts of labeled target data with labeled source and unlabeled target data [16]. Similar to unsupervised domain adaptation (UDA), SSDA methods for classification are also challenging to apply directly to regression tasks because they often rely on entropy measures from class prediction probabilities, which do not align well with continuous regression outputs.

The earliest proposed semi-supervised domain adaptation regression (SSDAR) method [17] aligned statistical moments and used graph Laplacians for unlabeled data, while recent method [18] used domain-specific regressors for invariant risks and adversarial learning to prevent the model from distinguishing between source and target distributions. Although these methods can be applied to some extent without specific regression assumptions, they rely on classification-based approaches, which may limit their effectiveness in addressing the challenges of regression tasks. To address these challenges, it is crucial to develop a method that directly considers the intrinsic properties of regression, rather than simply adapting classification techniques.

Fig. 1 illustrates the challenges posed by domain shifts in SSDAR, highlighting the distributional gap between source and target domains. To tackle these challenges, we introduce semi-supervised domain adaptation regression through whitening transform and dual consistency (WhiteCon). WhiteCon effectively reduces domain shift and addresses the unique challenges of regression tasks by combining domain-specific whitening transform (DWT) and dual consistency regularization.

The first component, DWT, eliminates correlations between features by transforming the feature covariance matrix into an identity matrix. Building upon the success of DWT in classification tasks [19], to our knowledge we are the first to apply it to regression tasks, providing a mathematical justification under ordinary least squares (OLS) assumptions. Unlike classification where DWT only aids feature alignment, in regression DWT improves model performance through regressor parameter variance reduction. Second, we introduce dual consistency regularization, which enforces consistency across both predictions and feature representations. Specifically, the variance consistency regularization enforces variance alignment among features extracted from weakly, strongly, and mixup-augmented samples of unlabeled target data. By matching the variances of strong and mixup augmentations to those of weak augmentations, the model achieves consistency across augmentation intensities and improves robustness. In summary, our method combines the variance-reducing effect of DWT with the robustness provided by dual consistency regularization, creating an effective method for SSDAR. The contributions of this study are summarized as follows:

(1) To the best of our knowledge, this study presents the first application of DWT to regression tasks with mathematical justification under the OLS framework. We demonstrate that DWT reduces the variance of regression parameters, thus improving task performance and providing theoretical support for the effectiveness of the proposed method.

(2) We introduce feature variance consistency regularization that aligns feature variances across different augmentations to improve robustness in SSDAR. This variance alignment promotes consistent feature distributions, reducing sensitivity to perturbations and enhancing generalization. Unlike existing methods that rely on complex formulations or focus on only one aspect independently, our approach is simple and addresses both feature and prediction consistency simultaneously in an intuitive manner.

(3) We propose WhiteCon, a simple yet effective framework that combines DWT and dual consistency regularization for SSDAR. Our method achieves state-of-the-art performance on various benchmark datasets and demonstrates its superiority through extensive evaluations, including ablation studies, regressor parameter variance analysis, varying proportions of labeled target data, and feature visualizations.

## 2 Related work

### 2.1 Domain Adaptation for Classification

UDA has been widely studied, particularly for classification tasks. One common approach, moment matching, aims to reduce distributional discrepancies by aligning the statistical moments of distributions [7, 8]. Maximum mean discrepancy (MMD) [20], a prominent technique, measures divergence by comparing their features in a reproducing kernel Hilbert space. However, MMD focuses on aligning input feature distributions but may overlook the complex dependencies between input features and continuous targets in regression tasks.

Another approach is adversarial learning, inspired by generative adversarial networks [21], which produces domain-invariant representations by training features that are difficult to distinguish between domains [9]. This method minimizes distributional differences through a competitive process with a domain discriminator [10]. Domain-adversarial neural networks (DANN) [6] align distributions through a gradient reversal layer. However, when domain discrepancies are substantial or highly complex, domain-invariant representations alone may not suffice to reduce the gap effectively.

The availability of labeled target data has increased interest in SSDA [22]. Saito et al. [13] demonstrated that UDA methods often struggled in SSDA settings. Various semi-supervised methods have applied consistency regularization, which aligns model predictions across augmented views of the same data to enhance stability and performance [23]. However, while effective for classification within SSDA, consistency regularization's effectiveness in regression remains unproven.

Most SSDA methods for classification rely on computing the entropy of class prediction probabilities using softmax [13, 24]. This reliance limits their applicability to regression tasks, which require models capable of handling continuous value predictions rather than class probability distributions.

### 2.2 Domain Adaptation for Regression

Cortes and Mohri [25] provided a theoretical analysis of domain adaptation for regression. Over time, several methods have been introduced for this task, often relying on importance weighting in non-deep learning models or focusing on creating invariant representations of features [18].

Recently, UDAR methods were proposed to address regression using only labeled source and unlabeled target data. Chen et al. [11] proposed representation subspace distance (RSD) for domain adaptation regression, which uses singular value decomposition to produce orthogonal bases while preserving feature scale. However, our empirical analysis reveals that the relationship between feature scale and performance was not always consistent. Nejjar et al. [14] proposed domain adaptation regression by aligning the inverse Gram matrices (DARE-GRAM) based on the closed-form OLS problem. However, deriving the pseudo-inverse matrix required specifying the low-rank property as a hyperparameter, potentially leading to excessive hyperparameter tun-ing. Wu et al. [26] proposed distribution-informed neural networks that build distribution-aware relationships using neural tangent kernel theory, but assume infinite-width networks that may not hold in practice. Dhaini et al. [27] proposed dictionary learning for subspace mapping, though performance depends on dictionary size and initialization.

SSDAR methods aim to address realistic scenarios where a small amount of labeled target data is available for model training. Singh and Chakraborty [17] proposed deep domain adaptation for regression (DeepDAR), the first SSDAR method using deep learning. They used MMD for distribution alignment and graph Laplacians for unlabeled target data. However, MMD only aligns feature distributions without ensuring consistent regression outputs, creating a weak connection to regression performance. Li et al. [18] proposed learning invariant representations and risks (LIRR) for both regression and classification, using domain-specific predictors and DANN. While theoretically grounded, LIRR's joint optimization prevents it from addressing regression-specific parameter stability under domain shift.

# 3 Proposed Methods

## 3.1 Preliminaries

Roy et al. [19] proposed DWT for reducing the distribution gap between source and target domains by eliminating feature correlations in unsupervised domain adaptation classification (UDAC). Unlike batch normalization, DWT applies batch whitening to transform the feature covariance matrix into an identity matrix. This ensures that features from both domains are mapped into a spherical distribution, which helps the model generalize across both domains. Let $Z$ represent the original feature vectors, and $\hat{Z}$ denote the decorrelated feature by applying a whitening matrix $W$. The whitening transformation is as follows:

$$\hat{Z} = W(Z - \bar{Z}), \tag{1}$$

where $\bar{Z}$ is the mean vector of the features, and $W$ is computed from the covariance matrix $\Sigma = (Z - \bar{Z})(Z - \bar{Z})^T$. Using the Cholesky decomposition, the covariance matrix $\Sigma$ is decomposed as $\Sigma = LL^T$, where $L$ is a lower triangular matrix. The whitening matrix is then obtained as $W = L^{-1}$. By applying $W$, the covariance matrix of the whitened feature $\hat{Z}$ becomes the identity matrix $I$, as shown below:

$$\mathrm{Cov}(\hat{Z}) = \hat{Z}\hat{Z}^T = W(Z - \bar{Z})(Z - \bar{Z})^T W^T = W\Sigma W^T = L^{-1}LL^T(L^T)^{-1} = I. \tag{2}$$

Consequently, DWT removes correlations between features and aligns the distributions between the source and target domains, improving generalization and reducing the impact of domain shifts.

Let $S = \{(x_i^s, y_i^s)\}_{i=1}^n$ represent a set of labeled data from the source domain, where $y_i^s$ denotes the continuous values associated with samples $x_i^s$. Similarly, let $T_{ul} = \{(x_i^{tul})\}_{i=1}^k$ represent a set of unlabeled data from the target domain. Additionally, let $T_l = \{(x_i^{tl}, y_i^{tl})\}_{i=1}^{m}$ represent a small set of labeled data in the target domain, where $m \ll n$ and typically $m \ll k$, indicating that only a small amount of target data is labeled. A key challenge is domain shift, where the source domain distribution $P(x^s)$ differs from the target domain distribution $P\left(x^{(tl, tul)}\right)$, i.e., $P(x^s) \neq P\left(x^{(tl, tul)}\right)$. This domain shift complicates model performance in the target domain, necessitating adaptation to address the distribution gap.

Given the data, the feature extractor $f(\cdot)$ transforms the source and target samples into features. The regressor $g(\cdot)$, implemented as a fully connected layer, uses these features to generate predictions for both domains, denoted as $\hat{y}^s = g(f(x^s))$ and $\hat{y}^{tl} = g(f(x^{tl}))$. To optimize the model using the labeled samples from both the source and target domains, the supervised loss $L_{sup}$ is defined as follows:

$$L_{sup} = \frac{1}{2}\left(\frac{1}{n}\sum_{i=1}^{n}(y_i^s - \hat{y}_i^s)^2 + \frac{1}{m}\sum_{i=1}^{m}\left(y_i^{tl} - \hat{y}_i^{tl}\right)^2\right), \tag{3}$$

where $n$ and $m$ represent the numbers of labeled samples in the source and target domains, respectively.

### 3.2 Motivations

Once features $Z$ are extracted, predicting the target $Y$ using regressor parameters $\beta$ can be formulated as an OLS problem [28]:

$$Y = Z\beta + \epsilon, \tag{4}$$

where $\epsilon$ is a random error. We construct our regressor $g(\cdot)$ as a single linear layer. This simplifies the mapping from features to predictions, clarifying the OLS problem. Under the OLS assumption, the variance of the parameter $\beta$ can be expressed as follows:

$$\mathrm{Var}\left(\hat{\beta}\right) = \sigma^2 (Z^T Z)^{-1}, \tag{5}$$

where $\sigma^2$ represents the variance of the error $\epsilon$. As shown in Equation (2), when features $Z$ are whitened using DWT in the feature extractor, the equation is modified as follows:

$$\mathrm{Var}\left(\hat{\beta}\right) = \sigma^2 \left(\hat{Z}^T \hat{Z}\right)^{-1} = \sigma^2 I, \tag{6}$$

where $\sigma^2 I$ indicates that the variance of the regressor parameters is reduced. This reduction in parameter variance can help improve the performance of the regression tasks [29]. These equations are applicable to SSDAR where a small amount of labeled target data is available for model training. Furthermore, a previous study [11] has shown that batch normalization negatively impacts domain adaptation regression, with better performance observed when it is disabled.

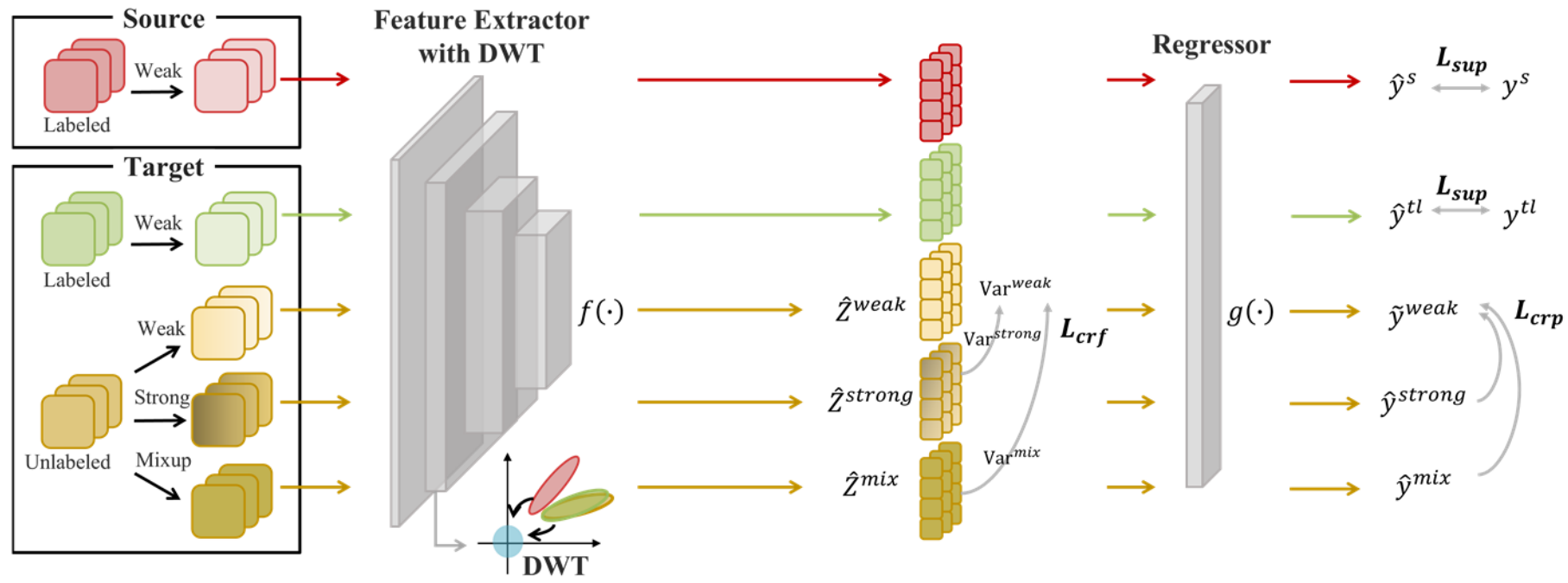

**Fig. 2.** Overview of the proposed framework, WhiteCon. DWT removes correlations between features and reduces the variance of regression model parameters. The proposed loss $L_{crf}$ trains the model to ensure that the variances of strongly and mixup-augmented features (Var$^{strong}$ and Var$^{mix}$) align more closely with the variance of weakly augmented features Var$^{weak}$, introducing consistency regularization for the augmented unlabeled features and enhancing model robustness.

### 3.3 Overview of the Proposed Method

Our method combines DWT and dual consistency regularization loss $L_{dcr}$ to address SSDAR. $L_{dcr}$ consists of prediction consistency regularization loss $L_{crp}$ and feature variance consistency regularization $L_{crf}$. The total loss $L_{total}$ is given by:

$$L_{total} = L_{sup} + L_{dcr} = L_{sup} + L_{crp} + L_{crf} . \tag{7}$$

Fig. 2 presents an overview of our proposed framework, WhiteCon. The framework takes source and target samples and applies weak, strong, and mixup augmentations to the unlabeled target samples. These augmented samples pass through a feature extractor with DWT, which removes feature correlations and stabilizes training by reducing the variance of regressor parameters. The loss $L_{crf}$ aligns the variances of strongly and mixup-augmented features with those of weakly augmented features, while $L_{crp}$ enforces consistency across predictions from these augmentations.

### 3.4 Dual Consistency Regularization

As discussed in Section 2.1, consistency regularization has proven effective in domain adaptation. Inspired by the semi-supervised regression approach of Sim et al. [30], we apply weak, strong, and mixup augmentations to unlabeled target samples $x^{tul}$ to obtain weak $x^{weak}$, strong $x^{strong}$ and mixup $x^{mix}$ augmented unlabeled target samples. Here, $x^{mix}$ is a linear combination of weak and strong augmented samples.

**Prediction Consistency Regularization.** After applying these augmentations, the feature extractor $f(\cdot)$, which includes DWT, transforms the augmented unlabeled target samples into feature representations $\hat{Z}^{weak}$, $\hat{Z}^{strong}$ and $\hat{Z}^{mix}$, respectively. Each of these feature representations belongs to $\mathbb{R}^{k\times d}$, where $d$ is the feature dimensionality. Using the regressor $g(\cdot)$, we obtain predictions $\tilde{y}^{weak}$, $\hat{y}^{strong}$ and $\hat{y}^{mix}$. We train the model to make the strongly augmented prediction $\hat{y}^{strong}$ and the mixup prediction $\hat{y}^{mix}$ close to the weakly augmented prediction $\tilde{y}^{weak}$ as the pseudo-label. The prediction consistency regularization loss $L_{cp}$ is defined as follows:

$$L_{cp} = \lambda_1 \left(\frac{1}{k}\sum_{i=1}^{k}\left(\tilde{y}_i^{weak} - \hat{y}_i^{strong}\right)^2 + \frac{1}{k}\sum_{i=1}^{k}\left(\tilde{y}_i^{weak} - \hat{y}_i^{mix}\right)^2\right), \tag{8}$$

where $\lambda_1$ is a hyperparameter for $L_{cp}$. Incorporating elements of a semi-supervised regression approach, $L_{cp}$ promotes consistent predictions across different augmentations, which is important in SSDAR settings with limited labeled target data.

**Feature Variance Consistency Regularization.** Unlike classification tasks, regression tasks predict continuous values for unlabeled data, making it challenging to apply threshold-based functions for pseudo-label refinement as in classification [31]. Therefore, introducing additional consistency beyond prediction consistency regularization is critical [30]. Previous studies proposed feature consistency across augmented unlabeled targets through complex calculations, such as simultaneously regulating variance, invariance, and covariance, or contrastive learning [32]. To simplify and enhance consistency regularization, we use feature variance from augmentations. The variance of the weakly augmented features $\mathrm{Var}^{weak}$ is calculated as follows:

$$\mathrm{Var}^{weak} = \frac{1}{k}\sum_{i=1}^{k}\left(\hat{Z}_i^{weak} - \mu^{weak}\right)^2, \tag{9}$$

where $\hat{Z}^{weak} \in \mathbb{R}^{k\times d}$ represents the weakly augmented features across $k$ samples and $d$ feature dimensions, and $\mu^{weak} \in \mathbb{R}^{1\times d}$ is the mean vector of these features. The variance is computed along each feature dimension, resulting in a variance vector with the same dimensionality $d$ as the feature matrix. Similar to $\mathrm{Var}^{weak}$ in Equation (9), the variances for strongly and mixup-augmented features ($\mathrm{Var}^{strong}$ and $\mathrm{Var}^{mix}$) are also computed along each feature dimension. The model is trained to align the variances of the strongly $\mathrm{Var}^{strong}$ and mixup-augmented features $\mathrm{Var}^{mix}$ with that of the weakly augmented features $\mathrm{Var}^{weak}$. The feature consistency regularization loss $L_{cf}$ is expressed as follows:

$$L_{cf} = \lambda_2 \left(\frac{1}{d}\sum_{j=1}^{d}\left|\mathrm{Var}_j^{weak} - \mathrm{Var}_j^{strong}\right| + \frac{1}{d}\sum_{j=1}^{d}\left|\mathrm{Var}_j^{weak} - \mathrm{Var}_j^{mix}\right|\right), \tag{10}$$

where $\lambda_2$ is a hyperparameter for $L_{cf}$. This variance alignment promotes consistent feature distributions across augmentations, reducing sensitivity to perturbations and enhancing generalization. Mean absolute error (MAE) is used for $L_{cf}$ because it handles differences evenly and prevents large errors from disproportionately affecting the training, leading to more stable variance alignment [33]. Unlike existing methods that focus on either prediction or feature consistency independently, our dual consistency regularization simultaneously addresses both aspects in a unified and straightforward manner.

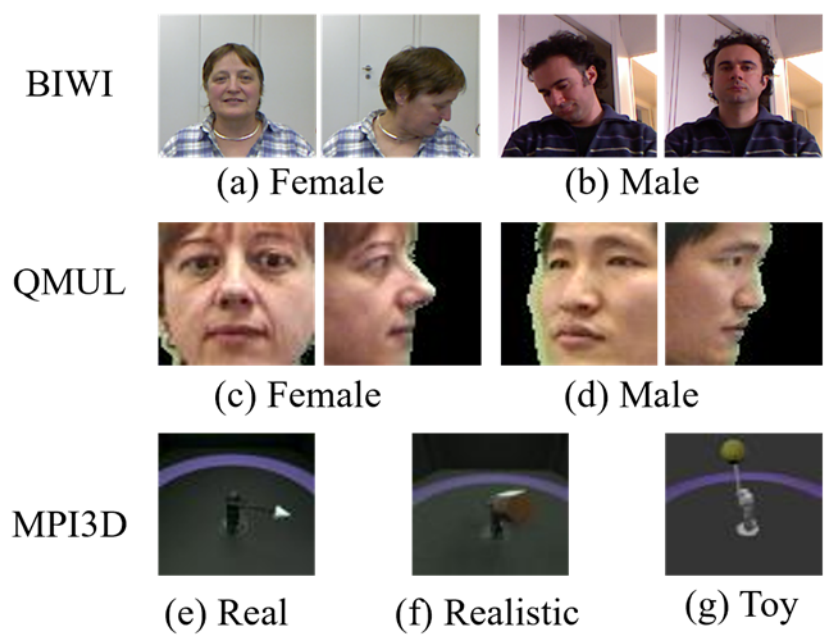

**Fig. 3.** Sample images from the three benchmark datasets, BIWI, QMUL, and MPI3D.

# 4 Experiments

## 4.1 Datasets

We used three benchmark datasets to address the SSDAR problem: BIWI [34] and QMUL [35] for head pose estimation, and MPI3D [36] for object position estimation. Sample images from each dataset are shown in Fig. 3. To satisfy the SSDAR setting, we randomly selected a small proportion (5%) of the target data as labeled target samples for training. Weak augmentation consisted of random cropping and Gaussian blur (excluding angle-altering transforms such as flips), and strong augmentation followed the FixMatch protocol.

**BIWI.** This dataset contains 5,874 images of 6 female (F) subjects and 9,804 images of 14 male (M) subjects, captured while turning their heads. We evaluated the dataset under two cases: M→F and F→M, predicting yaw, pitch, and roll angles.

**QMUL.** The dataset consists of 1,504 images of 8 female (F) subjects and 3,595 images of 40 male (M) subjects, also captured while turning their heads. We evaluated two cases: M→F and F→M, predicting yaw and pitch angles.

**MPI3D.** This dataset contains 1,036,800 3D object samples across three domains: real (RL), realistic (RC), and toy (T). The dataset has been used in UDAR settings [11, 14], and we adapted it for SSDAR, evaluating six cases: RL→RC, RL→T, RC→RL, RC→T, T→RL, and T→RC, predicting the vertical and horizontal axes.

## 4.2 Experimental Setup

**Implementation Details.** For the backbone model, we used ResNet50, which is widely used in domain adaptation regression and is suitable for the input image size [11, 14]. The source and target labels were scaled to the range [0, 1] using min-max normalization. The AdamW optimizer was used with a learning rate of 0.001 and a batch size of 48. The model was trained for 50 epochs, with a linear ramp-up for $L_{\sigma p}$ and $L_{\sigma f}$ over the first 20 epochs for stability. The optimal hyperparameters $\lambda_1$ and $\lambda_2$ were selected based on the highest validation score. An NVIDIA RTX 4090 GPU was used for all the experiments.

**Compared Methods.** We adapted the following methods as mentioned in Section 2 to the SSDAR setting for comparison: S+T, MMD, DANN, RSD, DARE-GRAM, DeepDAR, LIRR, and Full_T. S+T is a supervised learning method that only uses labeled source data and a small amount (5%) of labeled target data. Full_T is a model trained with the full set (100%) of labeled target data. We included UDAC methods (MMD and DANN) that can be applied to regression tasks, UDAR methods (RSD and DARE-GRAM), and SSDAR methods (DeepDAR and LIRR). Methods that could not be implemented under the same experimental conditions were excluded from the comparison [26, 27]. To apply UDA methods to the SSDAR setting, we incorporated $L_{sup}$ to train the model using the labeled target samples [18, 23].

### 4.3 Results

**BIWI.** Table 1 shows that WhiteCon achieved the best performance with an average $R^2$ of 0.86 and MAE of 4.71. Performance differences were more pronounced in F→M (WhiteCon: $R^2$ 0.87 vs. RSD: $R^2$ 0.80), while M→F results were comparable across methods ($R^2$: 0.83–0.85). WhiteCon outperformed the second-best method, RSD (average $R^2$: 0.82 (−0.04), MAE: 5.73 (+0.99)).
**QMUL.** WhiteCon achieved the best performance in both F→M and M→F cases, with an average $R^2$ of 0.83 and MAE of 9.35 (Table 1). Notably, most baseline methods showed performance similar to S+T (within 0.01 $R^2$ difference), suggesting saturation of existing approaches on QMUL. WhiteCon achieved meaningful gains over the second-best method, MMD (average $R^2$: 0.79 (−0.04), MAE: 10.52 (+1.17)).
**MPI3D.** As shown in Table 2, WhiteCon achieved the highest overall performance across all cases, with an average $R^2$ of 0.90 and MAE of 2.04. While scenarios with RL and RC as source domains achieved higher performance, the T→RL and T→RC cases proved challenging for comparative methods. Nevertheless, WhiteCon demonstrated robust performance even under these difficult conditions. The second-best method, DARE-GRAM (average $R^2$: 0.74 (−0.16), MAE: 3.78 (+1.74)), lagged substantially behind, representing the largest performance gap observed across the three datasets.

### 4.4 Analyses

**Ablation Study.** To evaluate the contribution of each component, we removed each module individually across three datasets (Table 3). While the most influential component varied across datasets, removing $L_{crf}$ (w/o $L_{crf}$) consistently led to the largest performance degradation, confirming its critical role in enhancing overall performance. Removing DWT (w/o DWT) or $L_{crp}$ (w/o $L_{crp}$) also led to performance drops of similar degree, indicating that both modules provide complementary contributions of comparable importance. These findings demonstrate that WhiteCon's superior performance relies on the integration of all three components, with $L_{crf}$ being the most important factor in general.

**Table 1.** Performance comparison of the proposed method and comparative methods on BIWI and QMUL in terms of $R^2$ and MAE. Result is reported as mean ± standard deviation with three different random seeds. **Bold** and underline indicate the best and the second-best results. The Wilcoxon rank-sum test is used to verify significant differences between the best and second-best average results, and is noted by *p*-value (*: *p*-value < 0.05).

| Method | BIWI | | | | | | QMUL | | | | | |
|---|---|---|---|---|---|---|---|---|---|---|---|---|
| | F→M | | M→F | | Average | | F→M | | M→F | | Average | |
| | $R^2$ ↑ | MAE↓ | $R^2$ ↑ | MAE↓ | $R^2$ ↑ | MAE↓ | $R^2$ ↑ | MAE↓ | $R^2$ ↑ | MAE↓ | $R^2$ ↑ | MAE↓ |
| S+T | 0.79±0.03 | 5.57±0.09 | 0.83±0.01 | 6.43±0.24 | 0.81±0.03 | 6.00±0.47 | 0.77±0.01 | 11.60±0.29 | 0.79±0.01 | 10.11±0.49 | 0.78±0.01 | 10.86±0.85 |
| MMD | 0.80±0.01 | 5.51±0.20 | 0.83±0.00 | 6.29±0.06 | 0.82±0.02 | 5.90±0.42 | 0.78±0.01 | 11.38±0.56 | 0.80±0.01 | 9.66±0.14 | 0.79±0.01 | 10.52±0.95 |
| DANN | 0.80±0.02 | 5.70±0.28 | 0.84±0.01 | 5.95±0.09 | 0.82±0.02 | 5.83±0.24 | 0.78±0.01 | 11.30±0.11 | 0.80±0.02 | 9.96±0.23 | 0.79±0.02 | 10.63±0.70 |
| RSD | 0.79±0.01 | 5.68±0.07 | **0.85±0.01** | 5.79±0.21 | 0.82±0.03 | 5.73±0.16 | 0.78±0.00 | 11.27±0.10 | 0.79±0.01 | 10.30±0.61 | 0.79±0.01 | 10.79±0.65 |
| DARE-GRAM | 0.79±0.01 | 5.55±0.20 | 0.81±0.01 | 6.31±0.20 | 0.80±0.01 | 5.93±0.43 | 0.77±0.02 | 11.68±0.33 | 0.80±0.00 | 9.77±0.22 | 0.78±0.02 | 10.72±1.00 |
| DeepDAR | 0.77±0.02 | 6.04±0.33 | 0.83±0.01 | 6.33±0.17 | 0.80±0.03 | 6.19±0.30 | 0.77±0.02 | 11.35±0.36 | 0.79±0.02 | 10.33±0.02 | 0.78±0.02 | 10.84±0.57 |
| LIRR | 0.79±0.02 | 5.83±0.39 | 0.84±0.01 | 6.11±0.09 | 0.81±0.03 | 5.97±0.31 | 0.77±0.01 | 11.33±0.35 | 0.79±0.02 | 9.96±0.14 | 0.78±0.02 | 10.65±0.74 |
| WhiteCon (ours) | **0.87±0.01** | **4.07±0.13** | **0.85±0.01** | **5.35±0.07** | ***0.86±0.01** | ***4.71±0.65** | **0.82±0.01** | **9.74±0.40** | **0.83±0.01** | **8.95±0.07** | ***0.83±0.01** | ***9.35±0.49** |
| Full_T | 0.94±0.01 | 3.30±0.11 | 0.98±0.00 | 2.62±0.34 | 0.96±0.02 | 2.96±0.42 | 0.88±0.01 | 9.31±0.31 | 0.92±0.00 | 6.64±0.24 | 0.90±0.03 | 7.98±1.37 |

**Table 2.** Performance comparison of the proposed method and comparative methods on MPI3D in terms of $R^2$ and MAE. Result is reported as mean ± standard deviation across two regression outputs with three different random seeds. **Bold** and underline indicate the best and the second-best results. The Wilcoxon rank-sum test is used to verify significant differences between the best and second-best average results, and is noted by *p*-value (*: *p*-value < 0.05).

| Method | RL→RC | | RL→T | | RC→RL | | RC→T | | T→RL | | T→RC | | Average | |
|---|---|---|---|---|---|---|---|---|---|---|---|---|---|---|
| | $R^2$ ↑ | MAE↓ | $R^2$ ↑ | MAE↓ | $R^2$ ↑ | MAE↓ | $R^2$ ↑ | MAE↓ | $R^2$ ↑ | MAE↓ | $R^2$ ↑ | MAE↓ | $R^2$ ↑ | MAE↓ |
| S+T | 0.88±0.01 | 2.54±0.08 | 0.66±0.12 | 4.61±0.81 | 0.81±0.05 | 3.33±0.49 | 0.70±0.10 | 4.34±0.70 | -0.03±0.01 | 9.05±0.11 | 0.34±0.04 | 6.66±0.17 | 0.56±0.32 | 5.09±2.24 |
| MMD | 0.89±0.01 | 2.51±0.10 | 0.78±0.03 | 3.86±0.16 | 0.82±0.06 | 3.14±0.46 | 0.79±0.02 | 3.80±0.21 | 0.09±0.03 | 8.49±0.15 | 0.31±0.01 | 7.22±0.29 | 0.61±0.30 | 4.84±2.23 |
| DANN | 0.89±0.03 | 2.46±0.25 | 0.77±0.06 | 3.85±0.51 | 0.83±0.02 | 3.41±0.29 | 0.81±0.03 | 3.57±0.31 | 0.03±0.03 | 8.94±0.17 | 0.24±0.05 | 7.61±0.42 | 0.60±0.33 | 4.97±2.43 |
| RSD | 0.92±0.01 | 2.13±0.18 | 0.75±0.05 | 4.18±0.35 | 0.84±0.03 | 3.04±0.32 | 0.80±0.05 | 3.69±0.53 | 0.04±0.01 | 8.77±0.08 | 0.34±0.01 | 6.88±0.31 | 0.61±0.32 | 4.78±2.33 |
| DARE-GRAM | 0.93±0.01 | 2.14±0.11 | 0.89±0.01 | 2.81±0.08 | 0.90±0.01 | 2.42±0.08 | 0.90±0.01 | 2.60±0.14 | 0.13±0.02 | 8.28±0.26 | 0.69±0.01 | 4.44±0.11 | 0.74±0.28 | 3.78±2.15 |
| DeepDAR | 0.89±0.02 | 2.68±0.20 | 0.74±0.09 | 4.28±0.58 | 0.76±0.03 | 4.12±0.24 | 0.79±0.04 | 3.79±0.29 | 0.08±0.03 | 8.57±0.22 | 0.24±0.08 | 7.76±0.65 | 0.58±0.31 | 5.20±2.21 |
| LIRR | 0.90±0.03 | 2.39±0.30 | 0.70±0.14 | 4.41±0.90 | 0.83±0.05 | 3.22±0.37 | 0.79±0.05 | 3.85±0.45 | 0.04±0.00 | 8.74±0.12 | 0.36±0.04 | 7.00±0.31 | 0.60±0.31 | 4.94±2.27 |
| WhiteCon (ours) | **0.98±0.00** | **1.06±0.04** | **0.95±0.01** | **1.93±0.15** | **0.98±0.00** | **1.28±0.09** | **0.95±0.00** | **1.83±0.05** | **0.61±0.06** | **4.32±0.45** | **0.95±0.01** | **1.82±0.14** | ***0.90±0.13** | ***2.04±1.09** |
| Full_T | 0.98±0.00 | 1.27±0.03 | 0.98±0.00 | 1.33±0.12 | 0.98±0.00 | 1.21±0.09 | 0.98±0.00 | 1.33±0.12 | 0.98±0.00 | 1.21±0.09 | 0.98±0.00 | 1.27±0.03 | 0.98±0.00 | 1.27±0.10 |

**Table 3.** Contribution of each component (DWT, $L_{crp}$ , $L_{crf}$ ) to the overall performance of WhiteCon on BIWI, QMUL, and MPI3D datasets in terms of MAE. Result is reported as mean ± standard deviation with three different random seeds. **Bold** indicates the component with the largest contribution to performance on each dataset, and underline denotes the second most influential component.

| Method | DWT | $L_{crp}$ | $L_{crf}$ | BIWI Average MAE | QMUL Average MAE | MPI3D Average MAE |
|---|---|---|---|---|---|---|
| Proposed | ✓ | ✓ | ✓ | 4.71±0.65 | 9.35±0.49 | 2.04±1.09 |
| w/o DWT | | ✓ | ✓ | 4.83±0.71 | 9.97±0.39 | 2.63±2.20 |
| w/o $L_{crp}$ | ✓ | | ✓ | 4.94±0.56 | 9.54±0.71 | **3.58±1.89** |
| w/o $L_{crf}$ | ✓ | ✓ | | **5.01±0.53** | **10.69±0.76** | 2.32±1.34 |

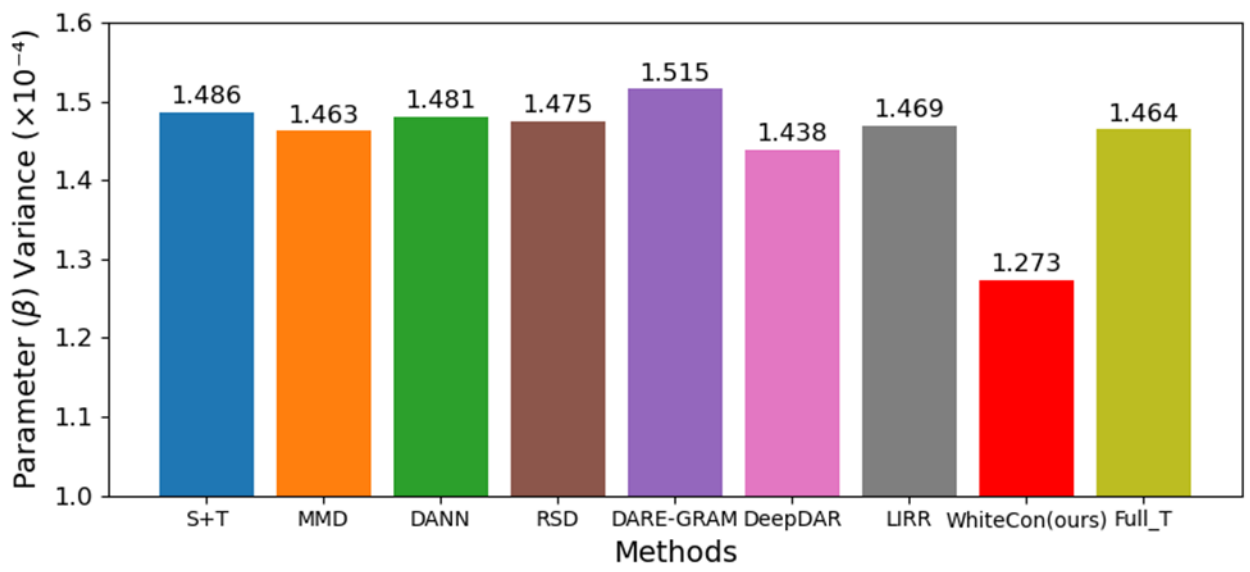

**Fig. 4.** Comparison of regressor parameter variance for QMUL F→M. The $x$-axis represents the different methods, the $y$-axis shows parameter variance values.

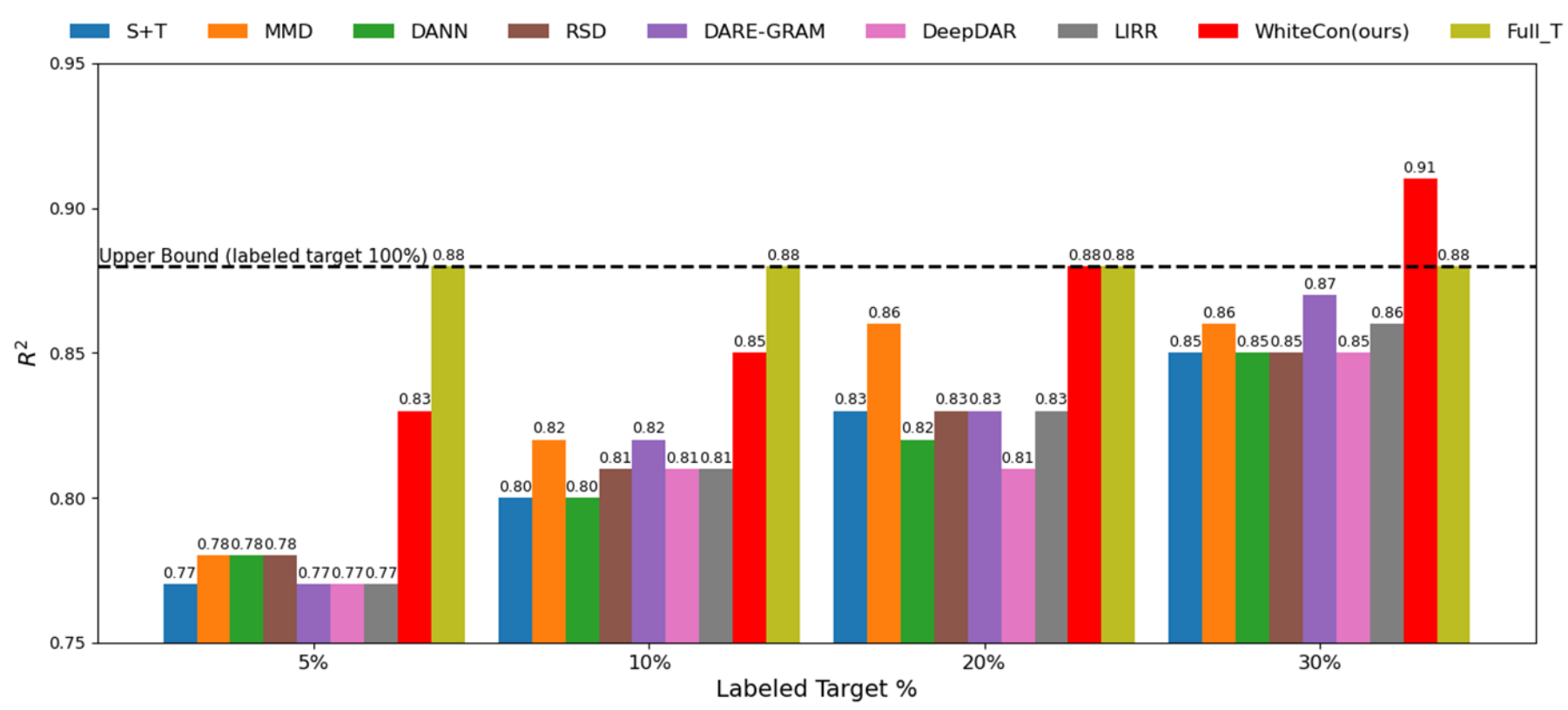

**Fig. 5.** Performance comparison with increasing proportions of labeled target data on QMUL F→M. The $x$-axis shows the proportions of labeled target data; the $y$-axis shows $R^2$. The dashed line indicates the upper bound (Full_T) with 100% labeled target data.

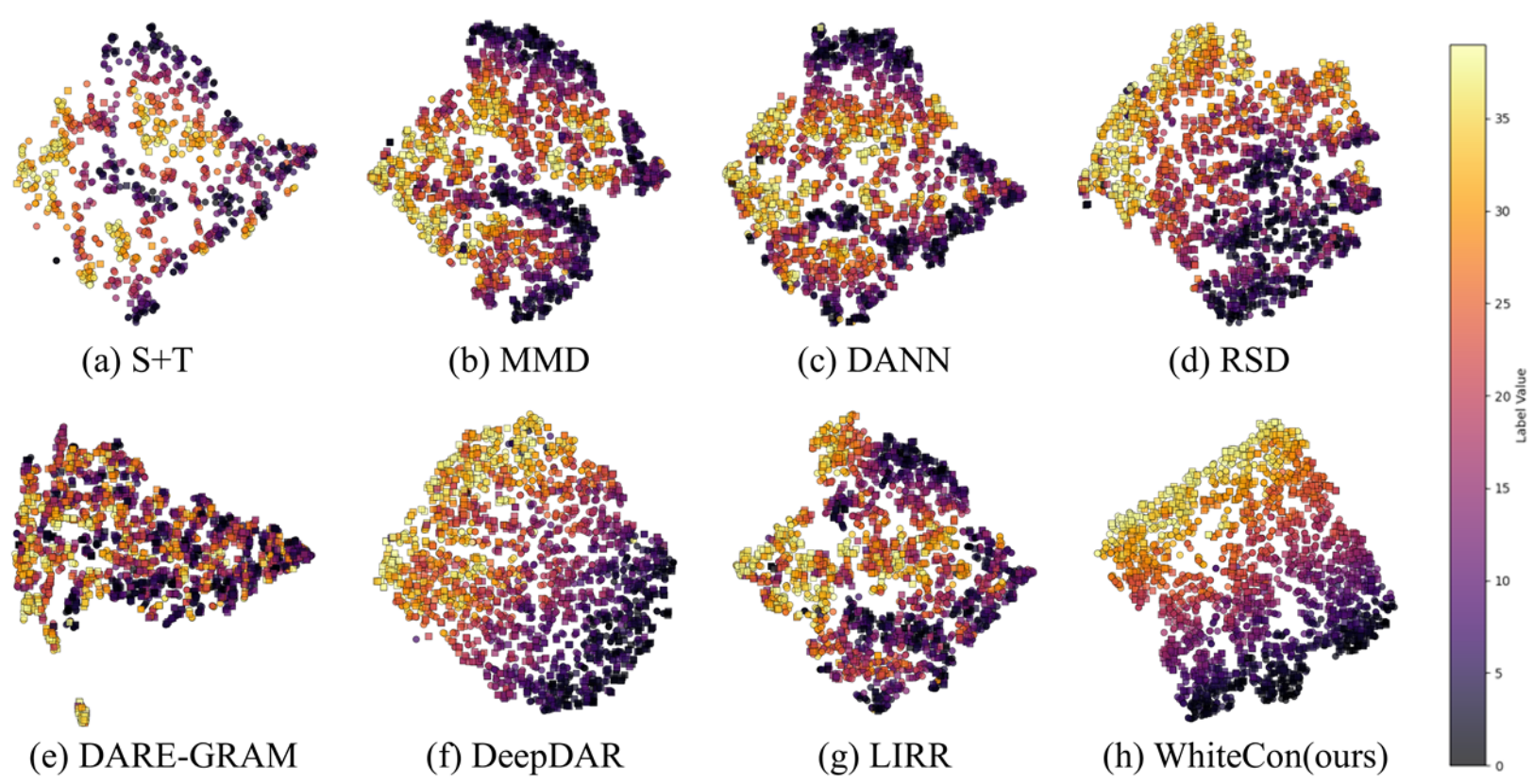

**Fig. 6.** Feature visualization results of comparison methods on MPI3D RL→RC. The color bar represents label values.

**Variance of Regressor Parameters.** To empirically validate the mathematical justification presented in Section 3.2, we conducted experiments to measure parameter variance across methods (Fig. 4). WhiteCon achieved the lowest variance of $1.273\times10^{-4}$, compared to $1.438\times10^{-4}$–$1.515\times10^{-4}$ for comparison methods. These results are consistent with Section 3.2, suggesting that lower parameter variance contributes to improved regression performance. Notably, while Full_T achieved high performance despite higher variance due to the absence of domain shift, among domain adaptation methods, WhiteCon both minimized variance and achieved the highest performance.
**Increasing Proportions of Labeled Target.** To verify WhiteCon's robustness across varying amounts of labeled target data, we conducted experiments with 5%, 10%, 20%, and 30% proportions. Fig. 5 shows that WhiteCon consistently outperformed comparison methods across all proportions on QMUL F→M. Remarkably, WhiteCon with 30% labeled data ($R^2$ 0.91) exceeded Full_T trained on 100% data ($R^2$ 0.88).
**Feature Visualization**. To examine feature alignment in the regression context, where effective alignment should show continuous patterns along label values, we visualized features (from the feature extractor $f(\cdot)$) on MPI3D RL→RC using t-distributed stochastic neighbor embedding in Fig. 6. (a) S+T showed features largely unaligned with label values. While (d) RSD and (f) DeepDAR demonstrated improved alignment, confusion between label values remained, with dark colors scattered among bright colors. In contrast, (h) WhiteCon showed the clearest alignment, effectively grouping features with similar label values and demonstrating its superior ability to reduce the domain gap.

## 5 Conclusion

In this study, we proposed WhiteCon, a method for addressing SSDAR problems, combining DWT and dual consistency regularization to reduce domain discrepancies. DWT improves model performance under OLS assumptions by transforming the feature covariance matrix into an identity matrix, thus reducing the variance of regression parameters. Variance consistency regularization aligns feature variances across augmentations, improving model robustness. Experimental results on multiple benchmark datasets demonstrate that WhiteCon achieves state-of-the-art performance, confirming its effectiveness. Future work will extend WhiteCon to underexplored modalities such as time series and tabular data, which are widely used in real-world regression applications but remain understudied in SSDAR, requiring the design of modality-specific augmentation and adaptation strategies.

### Acknowledgments

This research was supported by Brain Korea 21 FOUR, the Ministry of Science and ICT (MSIT) in Korea under the ITRC support program supervised by the Institute for Information Communication Technology Planning and Evaluation (IITP-2026-RS-2020-0-01749), and the National Research Foundation of Korea grant funded by the Korea government (RS-2022-00144190).

# Appendix for WhiteCon: Semi-Supervised Domain Adaptation Regression Through Whitening Transform and Dual Consistency

Se Jin Sim[0009–0000–6028–1690] and Seoung Bum Kim[0000–0002–2205–8516]

School of Industrial and Management Engineering
Korea University, Seoul, Republic of Korea
{ssj259, sbkim1}@korea.ac.kr

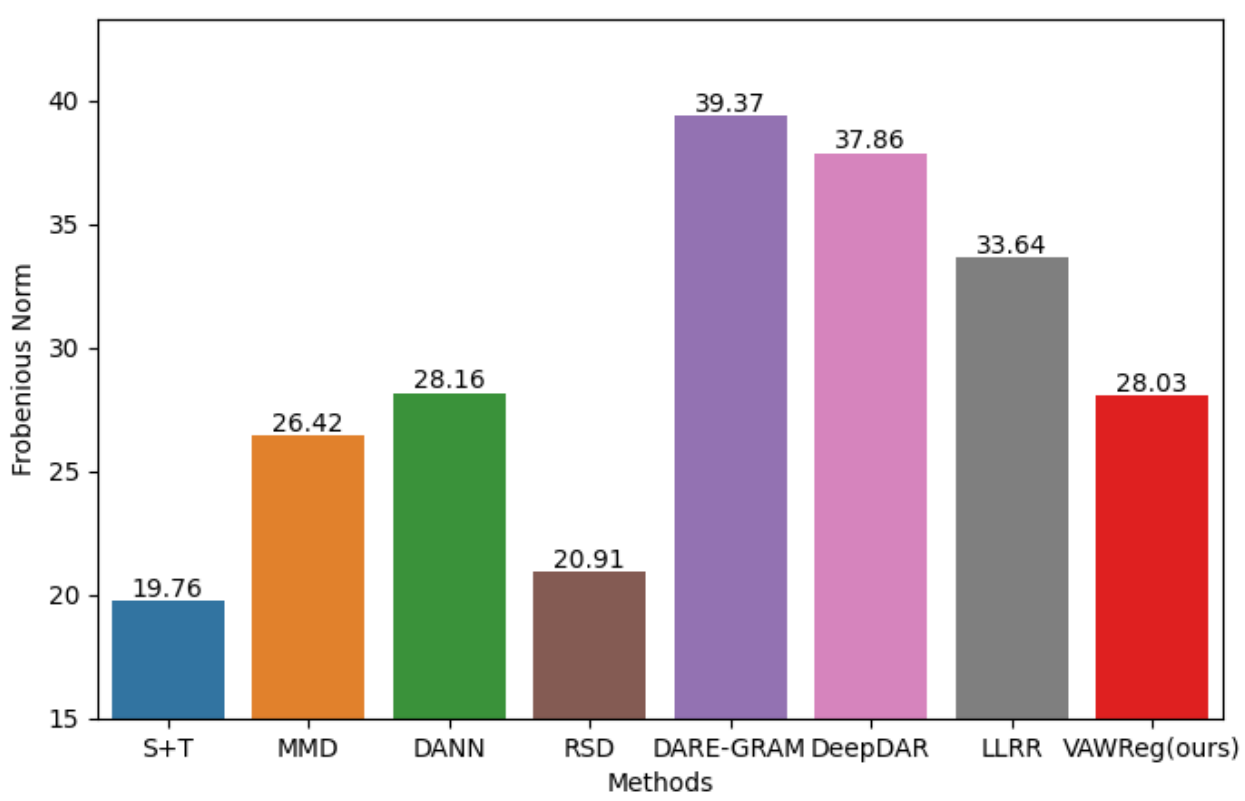

**Fig. A1.** Feature scale comparison across methods on BIWI F→M. The $x$-axis represents the different methods and the $y$-axis shows Frobenius norm values

## *A-1. Feature Scale.*

RSD argued based on experiments that maintaining a feature scale similar to supervised learning before domain adaptation yields optimal performance in domain adaptation regression. To investigate RSD's assertion about the importance of feature scale in domain adaptation regression, we conducted feature scale analysis using Frobenius norm. The Frobenius norm, an extension of the L2 norm for matrices, quantifies the scale of feature representations by taking the square root of the sum of the squared elements across all entries. Fig. A1 presents a feature scale comparison of methods on BIWI F→M. Although WhiteCon achieved the highest overall performance (as shown in Table 1), RSD showed a feature scale of 20.91, which is closest to that of S+T at 19.76. Additionally, MMD, which achieved the second-highest performance, had the second-largest feature scale of 26.42 among the methods. This result contrasts with RSD’s assertion that optimal performance in domain adaptation regression is achieved when the feature scale is close to S+T, suggesting that a similar feature scale is not a strict requirement for achieving high performance.

**Table A1.** Training (source, labeled target, and unlabeled target), validation, and testing data splits across three benchmark datasets.

| Dataset | Case | Source | Target | | | |
|---|---|---|---|---|---|---|
| | | | Labeled ($K$%) | Unlabeled | Validation | Testing |
| BIWI | F→M | 3,447 | 50 (5%) | 2,500 | 100 | 2,131 |
| | | | 100 (10%) | | | |
| | | | 200 (20%) | | | |
| | | | 300 (30%) | | | |
| | | | 1,000 (100%) | | | |
| | M→F | 5,624 | 35 (5%) | 1,500 | 70 | 1,247 |
| | | | 70 (10%) | | | |
| | | | 140 (20%) | | | |
| | | | 210 (30%) | | | |
| | | | 700 (100%) | | | |
| QMUL | F→M | 1,504 | 44 (5%) | 1,349 | 89 | 1,215 |
| | | | 89 (10%) | | | |
| | | | 179 (20%) | | | |
| | | | 268 (30%) | | | |
| | | | 896 (100%) | | | |
| | M→F | 2,500 | 15 (5%) | 600 | 30 | 569 |
| | | | 30 (10%) | | | |
| | | | 61 (20%) | | | |
| | | | 91 (30%) | | | |
| | | | 305 (100%) | | | |
| MPI3D | RL → RC | 3,000 | 50 (5%) | 2,000 | 100 | 10,437 |
| | RL → T | | 100 (10%) | | | 10,348 |
| | RC → RL | | 200 (20%) | | | 10,405 |
| | RC → T | | 300 (30%) | | | 10,348 |
| | T → RL | | 1,000 (100%) | | | 10,405 |
| | T → RC | | | | | 10,437 |

## *A-2. Datasets Settings*

We used three benchmark datasets: BIWI and QMUL for head pose estimation, and MPI3D for object position estimation, under SSDAR settings. Table A1 shows the number of training, validation, and testing samples for each dataset in each case. We used 10% of the labeled target data for validation, which was based on SSDA methods for classification [1-3]. To simulate different proportions of labeled target data, denoted as K%, we used values including 5%, 10%, 20%, and 30%, as shown in Fig. 5.

**Table A2.** Details of hyperparameter search for each method. Optimal values were selected based on validation performance.

| MMD | |
|---|---|
| Coefficient of MMD loss | {0.0005, 0.0001, 0.001, 0.01} |
| DANN | |
| Coefficient of DANN loss | {0.001, 0.01, 0.1, 1} |
| RSD | |
| Coefficient of RSD loss | {0.0001, 0.001} |
| Coefficient of bases mismatch penalization loss | {0.001, 0.01, 0.1} |
| DARE-GRAM | |
| Coefficient of angle loss | {0.001, 0.005, 0.05, 0.5, 0.1, 1} |
| Coefficient of trade-off loss | {0.00001, 0.0001, 0.0005, 0.001, 0.01} |
| DeepDAR | |
| Coefficient of MMD loss | {0.001, 0.01, 0.1, 1, 10} |
| Coefficient of semi-supervised loss | {0.001, 0.01, 0.1, 1, 10} |
| LIRR | |
| Coefficient of DANN loss | {0.01, 10, 100} |
| WhiteCon (ours) | |
| Coefficient of prediction consistency regularization loss ($\lambda_1$) | {0.1, 0.2, 0.3, 0.4, 0.5} |
| Coefficient of feature variance consistency regularization loss ($\lambda_2$) | {0.1, 0.2, 0.3, 0.4, 0.5} |

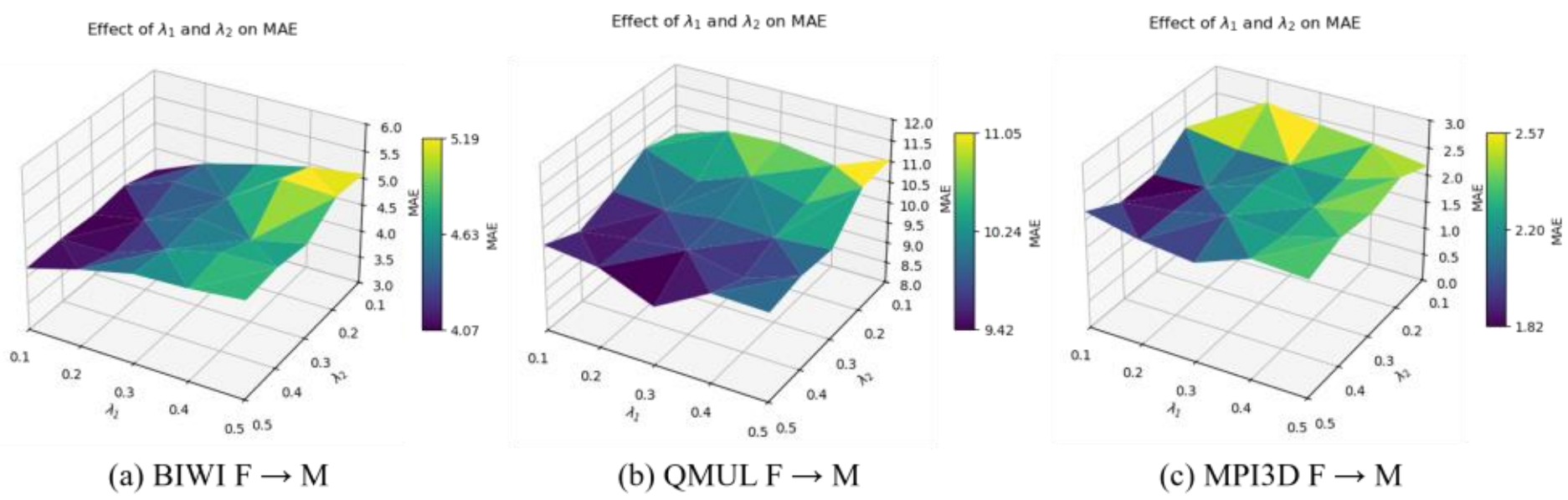

(a) BIWI F → M (b) QMUL F → M (c) MPI3D F → M

**Fig. A2.** MAE results from WhiteCon hyperparameter search across three benchmark datasets. The $x$-axis represents $\lambda_1$, the $y$-axis represents $\lambda_2$, and the $z$-axis represents MAE.

## *A-3. Hyperparameter Search*

We conducted extensive hyperparameter optimization for all compared methods to ensure fair evaluation. Table A2 details the hyperparameter ranges examined for each method, with optimal values selected based on validation performance. For WhiteCon, we systematically explored $\lambda_1$ and $\lambda_2$ values in the range {0.1, 0.2, 0.3, 0.4, 0.5}. Fig. A2 shows the hyperparameter search results for WhiteCon. We conducted experiments on a representative case from each of the three benchmark datasets to determine optimal parameters. Based on MAE performance, the optimal hyperparameters were identified as BIWI (0.1, 0.1), QMUL (0.3, 0.5), and MPI3D (0.1, 0.1).

## *A-4. Augmentation*

The images from all benchmarks were resized to 224 × 224 pixels. All comparison methods were trained using weakly augmented images. Following the augmentation guidelines for image angle prediction from Hu et al. [4], our weak augmentation consisted of random cropping and Gaussian blur, while excluding transformations that could alter the angle, such as flips. The strong augmentation strategy was identical to the one used in FixMatch [5], a widely-used semi-supervised method. As mentioned in Section 3.4, the mixup augmentation is a linear combination of weakly and strongly augmented samples.